\pdfoutput=1
\PassOptionsToPackage{table}{xcolor}
\documentclass[11pt]{article}

\usepackage[preprint]{acl}

\usepackage{times}
\usepackage{latexsym}
\usepackage[T1]{fontenc}
\usepackage[utf8]{inputenc}
\usepackage{microtype}
\usepackage{amsmath}
\usepackage{amssymb}
\usepackage{mathtools}
\usepackage{amsthm}
\usepackage{enumitem}
\usepackage{graphicx}
\usepackage{subcaption}
\usepackage{caption}
\usepackage{wrapfig}
\usepackage{pifont}
\usepackage{float}
\usepackage{booktabs}
\usepackage[most]{tcolorbox}
\usepackage{tabularx}
\usepackage{bm}
\usepackage{array}
\usepackage{multirow}

\definecolor{LightCyan}{rgb}{0.88,1,1}

\newcommand{\UCS}{\textsc{UCS}}

\title{\UCS: Estimating Unseen Coverage for Improved In-Context Learning}

\author{
  \textbf{Jiayi Xin\textsuperscript{1}},
  \textbf{Xiang Li\textsuperscript{1}},
  \textbf{Evan Qiang\textsuperscript{1}},
  \textbf{Weiqing He\textsuperscript{1}},
  \textbf{Tianqi Shang\textsuperscript{1}},
  \\
  \textbf{Weijie J.\ Su\textsuperscript{1}},
  \textbf{Qi Long\textsuperscript{1}}
\\
\\
  \textsuperscript{1}University of Pennsylvania
  \\
  \small{
    \textbf{Correspondence:} \href{mailto:jiayixin@seas.upenn.edu}{jiayixin@seas.upenn.edu}, \href{suw@wharton.upenn.edu}{suw@wharton.upenn.edu} and \href{mailto:qlong@upenn.edu}{qlong@upenn.edu}
  }
}

\begin{document}
\maketitle
\begin{abstract}
In-context learning (ICL) performance depends critically on which demonstrations are placed in the prompt, yet most existing selectors prioritize heuristic notions of relevance or diversity and provide limited insight into the \emph{coverage} of a demonstration set.
We propose \textbf{Unseen Coverage Selection (\UCS{})}, a training-free, subset-level coverage prior motivated by the principle that a good demonstration set should \emph{expose the model to latent cluster unrevealed by the currently selected subset}.
\UCS{} operationalizes this idea by \textbf{(1)} inducing discrete latent \emph{clusters} from model-consistent embeddings and \textbf{(2)} estimating the number of unrevealed clusters within a candidate subset via a Smoothed Good--Turing estimator from its empirical frequency spectrum.
Unlike previous selection methods, \UCS{} is coverage-based and training-free, and can be seamlessly combined with both query-dependent and query-independent selection baselines via a simple regularized objective.
Experiments on multiple intent-classification and reasoning benchmarks with frontier LLMs show that augmenting strong baselines with UCS consistently improves ICL accuracy by up to \textbf{2--6\%} under the same selection budget, while also yielding insights into task- and model-level latent cluster distributions. Code is available at {\small\url{https://github.com/Raina-Xin/UCS}}.
\end{abstract}
% It can be layered on top of both query-dependent (e.g., DPP, MDL) and query-independent (e.g., VoteK) baselines through a simple regularized objective.
% Across multiple intent classification benchmarks and frontier LLMs, augmenting strong selection baselines with \UKS{} yields consistent gains in ICL accuracy, improving performance by up to \textbf{X--Y\%} under the same selection budget, while also providing interpretable insights into task- and model-level knowledge distributions. Code is available at {\small\url{URL}}.

\section{Introduction}

In-context learning (ICL) enables large language models (LLMs) to perform new tasks by conditioning on a small set of input–output demonstrations, without parameter updates \citep{brown2020language}. While highly effective across tasks \citep{wei2022emergent, wang2024learning}, ICL is notoriously sensitive to the choice of demonstrations---under the same selection budget, different demonstration sets can lead to large performance variations \citep{su2022selective, min2022rethinking}. This sensitivity poses a central challenge in modern deployments, where ICL demonstrations are selected from large candidate pools under limited annotation budgets \citep{wang2025rdes}. 
Although a large body of work on automatic demonstration selection has improved ICL performance---via similarity-based \citep{ye2023compositional, su2022selective}, diversity-based \citep{luo2023dr, yang-etal-2023-representative}, and information-theoretic methods \citep{wu2023self, van2024context, qin-etal-2024-context, scarlatos2023reticl, wang2025rdes}---it remains unclear how much of the task’s underlying \emph{latent cluster space} is revealed by the chosen subset.

\begin{figure}[!htbp]  % You can also use [htbp] instead of [H] if you want more flexible positioning
\centering
\includegraphics[width=0.49\textwidth]{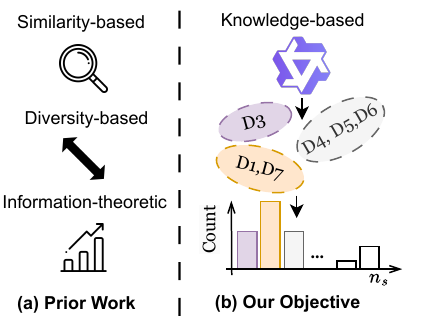}  % Change width and filename as needed
\vspace{-1em}
\caption{ Contrast to prior methods (\textbf{left}), our approach reasons at the subset level by clustering demonstrations into latent clusters and using their frequency to measure coverage (\textbf{right}).}

\label{fig:fig1_comparison}
\vspace{-1em}
\end{figure}

In this work, we argue that \emph{coverage} provides a complementary dimension for reasoning about demonstration selection (Figure \ref{fig:fig1_comparison}). Prior studies suggest that ICL performance depends not only on relevance to the test input, but also on exposure to diverse latent concepts and skills \citep{min2022rethinking, peng2024revisiting}. From our perspective, a demonstration is valuable precisely when it introduces \emph{previously unseen latent clusters}. Nevertheless, latent cluster coverage alone is insufficient to serve as a standalone criterion, since relevance- and diversity-based methods capture important instance-level signals. We therefore treat coverage as a \emph{prior} that biases demonstration selection toward subsets with broader latent cluster exposure, rather than as a replacement for existing objectives. 

To operationalize this idea, we introduce \textbf{Unseen Coverage Selection (\UCS{})}, a statistically principled framework for estimating latent cluster coverage in ICL demonstration selection. At a high level, \UCS{} entails two steps: it first clusters model embeddings of demonstrations to induce a set of discrete latent \emph{clusters}, and then applies the Smoothed Good--Turing estimator \citep{orlitsky2016optimal} to the resulting empirical frequency spectrum of a candidate subset to estimate how many additional clusters are likely to remain \emph{unrevealed}---that is, latent clusters not yet covered by the selected demonstration subset under the same candidate-pool sampling process. This estimate provides a subset-level measure of cluster coverage, which we use as a lightweight coverage prior that can be incorporated into existing selection objectives. As a result, \UCS{} is training-free, and it can be layered on top of both query-dependent and query-independent methods via a simple regularized objective with minimal computational overhead.

We evaluate \UCS{} across three intent classification benchmarks, three Big-Bench Extra Hard (BBEH) reasoning tasks, and three frontier LLMs. Our experiments show that augmenting strong baseline selectors with \UCS{} consistently improves ICL accuracy by up to \textbf{2--6\%} under the same selection budget. Beyond performance gains, \UCS{} provides interpretable insights into how latent clusters are distributed across tasks and models, revealing systematic differences in the prevalence of common versus rare clusters.

Our contributions are summarized as follows:
\begin{itemize}[leftmargin=*, itemsep=0pt, topsep=0pt]
% \vspace{-1em}
    \item[$\star$] We introduce \emph{coverage} as a complementary subset-level
    objective for ICL demonstration selection, focusing on estimating unseen
    latent clusters beyond instance-level relevance.
% \vspace{-1em}
    \item[$\star$] We propose \UCS{}, a lightweight, training-free coverage prior based on the
    Smoothed Good--Turing estimation that can be plugged into existing selection
    methods.
% \vspace{-1em}
    \item[$\star$] \UCS{} consistently improves ICL accuracy across three intent
    classification datasets, three BBEH reasoning tasks, and three LLM backbones, achieving gains of up to \textbf{6.2\%} for
    query-independent selection and stable improvements for strong query-dependent baselines
    under the same budget.
\end{itemize}

\section{Related Work}

\paragraph{In-Context Learning and Demonstration Selection.}

ICL performance is highly sensitive to the choice of demonstrations: under a fixed
budget, different selections can lead to large accuracy variations
\citep{su2022selective, min2022rethinking}. Early work addresses this problem using heuristic similarity or diversity criteria, selecting demonstrations that are either semantically close to the test query or diverse among themselves \citep{liu2022makes, lu2022fantastically, wu2023self, su2022selective, ye2023compositional, levy2023diverse, luo2023dr, agrawal2023context}. The OpenICL framework \citep{wu2023openicl} provides a unified evaluation of such strategies and includes representative methods based on similarity (e.g., VoteK) \citep{su2022selective}, diversity via determinantal point processes (DPP) \citep{yang-etal-2023-representative}, and information-theoretic criteria such as minimum description length (MDL) \citep{wu2023self}. However, these approaches do not explicitly characterize how much of the task’s
underlying \emph{latent space} is exposed by a selected demonstration subset.

\paragraph{Query-Adaptive and Iterative Retrieval.}

Several methods make demonstration selection query-adaptive by incorporating
relevance signals, iterative refinement, or learned policies.
Influence-based approaches such as InfICL estimate the effect of individual
training examples on predictions \citep{van2024context}, while iterative methods
(e.g., IDS) refine demonstrations using intermediate chain-of-thought signals
\citep{qin-etal-2024-context}.
These ideas are closely related to reasoning-based prompting techniques, including
chain-of-thought, zero-shot reasoning, self-consistency, and automatic CoT
construction \citep{wei2022chain, kojima2022large, wang2022self, zhang2022automatic}.
Other work formulates selection as a sequential decision problem, optimized via
reinforcement learning or query-conditioned mechanisms
\citep{scarlatos2023reticl, wang2025rdes, liu2022makes, an2023skill, liu2024se},
with recent extensions exploring alternative reward designs
\citep{zhang2025reward}.
These methods emphasize instance-level optimization but typically require
additional training or incur non-trivial inference overhead.

\paragraph{Subset-Level Coverage and Unseen Information.}
Beyond instance-level and query-adaptive strategies, prior work has examined
demonstration selection from a \emph{subset-level} perspective, emphasizing
latent coverage rather than relevance alone.
Empirical studies show that ICL benefits from exposing diverse concepts and skills,
even when they are not semantically closest to the test query
\citep{min2022rethinking, peng2024revisiting, levy2023diverse, gupta2023coverage,
ye2023complementary, an2023skill, lu2022fantastically, min2022metaicl}.
This view aligns with diversity- and representativeness-based objectives, such as
DPP and related subset selection methods, which trade off redundancy and coverage
under a fixed budget
\citep{yang-etal-2023-representative,
kulesza2012determinantal, lin2011class}.
However, most existing approaches lack a principled way to quantify how much latent
structure is covered or remains unseen in a selected set
\citep{wu2023self, qin-etal-2024-context}.
This limitation is closely related to classical unseen-species estimation
\citep{orlitsky2016optimal} and recent efforts to quantify unseen knowledge in LLM
evaluation (e.g., KnowSum) \citep{li2025evaluating}, which do not address ICL
demonstration selection under a fixed budget.

\section{Unseen Coverage Selection for In-Context Learning}
\label{sec:ucs}

\subsection{Preliminaries and Notation}
\label{sec:ucs_prelim}

\noindent \textbf{Problem setup.}

Let $\mathcal{D}=\{(x_i, y_i)\}_{i=1}^{N}$ denote a pool of labeled demonstrations, and let $B$ denote the selection budget. For each test input $x_{\text{test}}$, the goal is to select a subset $S \subseteq \mathcal{D}$ with $|S| = B$ and construct an in-context prompt by concatenating the labeled examples in $S$, followed by $x_{\text{test}}$.

\noindent \textbf{Base demonstration selectors.}
We view existing ICL demonstration selection methods as defining a subset utility function $U_{\text{base}}(S; x_{\text{test}})$. Some selectors are \emph{query-dependent}, which produce different subset for each test input (e.g., DPP and MDL), while others are \emph{query-independent} that select a single global subset shared across all test examples (e.g., VoteK).

\subsection{Algorithm Overview of \UCS{}}
\label{sec:ucs_overview}

Our key hypothesis is that strong demonstration sets should \emph{cover} a broad range of latent units in the pool, rather than over-sampling redundant regions. To operationalize this idea, we introduce \UCS{} (Unseen Coverage Selection) as a \emph{subset-level coverage prior} that can be \emph{layered on top of} existing demonstration selection methods. \UCS{} is \emph{not} used as a standalone retriever; instead, it regularizes the selection objective by biasing it toward subsets with higher expected coverage under a fixed selection budget.

\begin{figure*}[htbp] 
\centering
\includegraphics[width=1.0\textwidth]{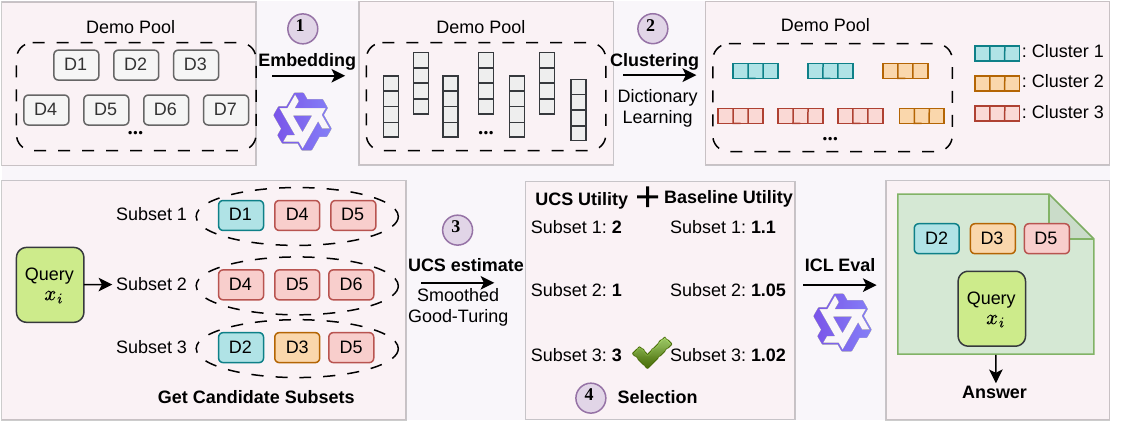}  % Change width and filename as needed
% \vspace{-1em}
\caption{Schematics of \UCS{}. \textbf{(1)} Demonstrations are embedded in a model-consistent space. \textbf{(2)} Embeddings are discretized into latent clusters via dictionary learning and clustering. \textbf{(3)} Candidate subsets are scored using an SGT estimation-regularized objective with a base selection utility.  \textbf{(4)} The selected subset is used for ICL.}
\label{fig:method}  % For referencing the figure in text with
\vspace{-1em}
\end{figure*}

Concretely, \UCS{} consists of three stages: (i) representing demonstrations in a model-consistent embedding space (Section~\ref{sec:ucs_embed}); (ii) inducing discrete latent clusters via dictionary learning and clustering (Section~\ref{sec:ucs_units}); and (iii) estimating subset-level coverage using a Smoothed Good--Turing estimator (Section~\ref{sec:ucs_coverage}). These components are combined with existing ICL selectors through a \UCS{}-regularized objective, which we instantiate for DPP, MDL, and VoteK in Section~\ref{sec:ucs_objective}. Figure~\ref{fig:method} illustrates the end-to-end workflow.

\subsection{Model-Consistent Representation}
%of Demonstrations}
\label{sec:ucs_embed}

\UCS{} is designed to be \emph{model-consistent}: the representation space used to define latent clusters should align with the LLM used at inference time for ICL. Accordingly, we embed the demonstration pool using the \emph{same} backbone model during evaluation, so that distances and clusters reflect the inductive biases of the model being prompted.

\noindent \textbf{Embedding procedure.}
For each demonstration $(x_i,y_i)\in\mathcal{D}$, we embed \emph{only the input} $x_i$ (excluding labels) to avoid leaking task-specific surface information into the representation.
Let $\mathbf{H}_i^{(\ell)}\in\mathbb{R}^{T_i\times d}$ denote the hidden states at layer $\ell$ for the tokenized input $x_i$, with an attention mask $\mathbf{m}_i\in\{0,1\}^{T_i}$ indicating non-padding tokens. We obtain a fixed-length embedding $\mathbf{e}_i\in\mathbb{R}^{d}$ via \textbf{masked mean pooling}:
\begin{equation}
\mathbf{e}_i
=
\frac{\sum_{t=1}^{T_i} m_{i,t}\,\mathbf{H}^{(\ell)}_{i,t}}
{\sum_{t=1}^{T_i} m_{i,t}}.
\label{eq:masked-mean-pooling}
\end{equation}
The resulting $\{\mathbf{e}_i\}_{i=1}^N$ serve as the continuous representation of the
demonstration pool for subsequent clustering and coverage estimation. Additional choices, including the layer $\ell$, alternative pooling strategies, dimensionality reduction, are deferred to Appendix~\ref{app:embedding_details}.

\subsection{Inducing Discrete Clusters from Continuous Embeddings}
\label{sec:ucs_units}

Smoothed Good--Turing–style unseen estimators operate on \emph{discrete objects}. To apply them to continuous LLM embeddings, we discretize each embedding into a \emph{cluster assignment} via a three-step procedure: (i) learning a latent dictionary over embeddings, (ii) encoding each example as a structured code over dictionary atoms, and (iii) clustering these codes to induce discrete units.

\paragraph{Dictionary learning as a latent basis.}
Let $E=[e_1;\dots;e_N]\in\mathbb{R}^{N\times d}$ denote the embedding matrix of the demonstration pool.
For numerical stability and computational efficiency, we optionally standardize features and apply PCA to obtain $\tilde{E}\in\mathbb{R}^{N\times d'}$, where $d'\ll d$ in our implementation.
We then learn a dictionary with $K$ atoms by approximating each embedding as
\begin{equation}
\tilde{e}_i \;\approx\; D r_i,\qquad r_i\in\mathbb{R}^{K},
\label{eq:dict}
\end{equation}
where $D\in\mathbb{R}^{d'\times K}$ is the learned basis and $r_i$ is the code for example $i$. We adopt \textbf{ridge coding} to avoid pathological sparsity and to preserve meaningful geometry in the code space. After fitting $D$, we compute codes via the ridge-regularized projection
\begin{equation}
r_i \;=\; \arg\min_{r\in\mathbb{R}^{K}} \;\|\tilde{e}_i - D r\|_2^2 + \alpha\|r\|_2^2,
\label{eq:ridge_code}
\end{equation}
which admits a closed-form solution and can be efficiently applied in batch.

\paragraph{Clustering in code space.}
A naive discretization strategy would assign each example to its highest-magnitude atom (i.e., $\arg\max_j |r_{ij}|$), but this ignores the inherently \emph{compositional} structure of the codes. Instead, we treat the full code vector as a latent \emph{cluster signature} and perform clustering directly in the code space. Because code norms can vary due to input length and embedding scale, we normalize each code before clustering:
\begin{equation}
\bar r_i \;=\; \frac{r_i}{\|r_i\|_2+\varepsilon}.
\label{eq:code_norm}
\end{equation}
We then apply DBSCAN \citep{ester1996density} to $\{\bar r_i\}_{i=1}^N$ using cosine distance, producing cluster assignments $c_i\in\{-1,0,1,2,\ldots\}$. Points labeled as noise ($c_i = -1$) are each assigned a unique singleton cluster.
After this remapping, we obtain $c_i' \in \{1,2,\ldots\}$, which serves as the \textbf{discrete cluster ID} required by the SGT estimator; each cluster serves as a discrete latent type for downstream coverage estimation.

\paragraph{Why dictionary learning plus clustering?}
Empirically, atom activations exhibit a heavy-tailed distribution: a small number of atoms are
frequently used, while many others are rare. As a result, argmax-based assignment over-emphasizes
frequent atoms and fails to capture multi-atom structures. In contrast, clustering in the code space
groups examples that share similar \emph{combinations} of atoms, yielding clusters that better
reflect recurring latent patterns while preserving a long tail of fine-grained units. Additional
discussion and implementation details, including the DBSCAN $\texttt{eps}$ heuristic and comparisons
with argmax discretization, are provided in Appendix~\ref{app:dbscan_eps} and
Appendix~\ref{app:dict_vs_argmax}.

\subsection{UCS Coverage Functional}
\label{sec:ucs_coverage}

\paragraph{Frequency spectrum of clusters.}
Let $S$ be a candidate demonstration subset and let $c_i$ denote the discrete cluster
assigned to example $i$ (Section~\ref{sec:ucs_units}).
We define the multiplicity of unit $u$ in $S$ by
\begin{equation}
n_u(S)
=
\sum_{i\in S}\mathbb{I}[c_i=u],
\end{equation}
and the corresponding frequency-of-frequencies (or \emph{spectrum}) by
\begin{equation}
f_s(S)
=
\big|\{u : n_u(S)=s\}\big|.
\end{equation}

\paragraph{Smoothed Good--Turing estimation.}
Given the frequency spectrum $\{f_s(S)\}_{s \ge 1}$, we apply a Smoothed Good--Turing (SGT) estimator \citep{good1953population, orlitsky2016optimal} to predict the number of \emph{new} clusters that would be observed if we were to draw $m$ additional samples from the same candidate pool (we define the expansion factor $t \coloneqq m/|S|$).
Let $\widehat{U}_t(S)$ denote this unseen estimate.
The classical Good--Turing functional is
\begin{equation}
\widehat{U}^{\mathrm{GT}}_{t}(S)
=
-\sum_{s=1}^{M} (-t)^s\, f_s(S),
\label{eq:gt_trunc}
\end{equation}
where we truncate the alternating series at a maximum count $M$ (default $M{=}20$) to reduce variance from sparse high-multiplicity bins.
To further stabilize extrapolation, we apply offset-damped weights $w_s(t,\alpha)\in[0,1]$ that discount higher-order terms:
\begin{equation}
\widehat{U}^{\mathrm{SGT}}_{t,M}(S)
=
-\sum_{s=1}^{M} (-t)^s\, w_s(t,\alpha)\, f_s(S),
\label{eq:sgt_weighted}
\end{equation}
with negative or non-finite values clamped to zero.
Full details on the weighting scheme and hyperparameters are in Appendix~\ref{app:ucs_estimator_details}.

\paragraph{UCS coverage functional.}
Instead of optimizing the unseen mass alone, we define the UCS coverage functional $\Phi_{\text{UCS}}(S) $ as the estimated total number of clusters, i.e., $\widehat{K}_t(S)$, in a set $S$:
\begin{equation}
\Phi_{\text{UCS}}(S) \coloneqq
\widehat{K}_t(S)
=
K_{\text{seen}}(S)
+
\widehat{U}_t(S),
\label{eq:khat}
\end{equation}
\begin{equation}
K_{\text{seen}}(S)
=
\big|\{u : n_u(S)>0\}\big|.
\label{eq:kseen}
\end{equation}

Maximizing $\Phi_{\text{UCS}}(S)$ encourages demonstration subsets that both cover many distinct clusters and exhibit frequency spectra indicative of additional unseen clusters.
For numerical robustness, negative or non-finite values of $\widehat{U}_t(S)$ are clamped to zero, and designated noise labels (e.g., DBSCAN noise) are excluded from the frequency spectrum.

\subsection{UCS-Regularized Selection}
\label{sec:ucs_objective}

UCS is not a standalone demonstration selector.
Instead, it defines a \emph{set-level coverage prior} that regularizes existing in-context learning selection objectives.
Given a budget $B$, we select demonstrations by solving
\begin{equation}
S^{*}
=
\arg\max_{|S|=B}
\Big(
U_{\text{base}}(S; x_{\text{test}})
+
\lambda\,\Phi_{\text{UCS}}(S)
\Big),
\label{eq:ucs_objective}
\end{equation}
where $U_{\text{base}}$ denotes the utility induced by a base selector (e.g., DPP, MDL, or VoteK), $\Phi_{\text{UCS}}(S)$ is the UCS coverage functional defined in Section~\ref{sec:ucs_coverage}, and $\lambda\ge 0$ controls the strength of the coverage regularization.

Crucially, $\Phi_{\text{UCS}}$ operates at the \emph{subset level}: it depends on the joint frequency spectrum induced by $S$ and does not decompose into per-example scores. As a result, it can be interpreted as an \emph{unnormalized prior} over subsets, biasing selection toward demonstration sets with broader and more extensible
coverage. Throughout this work, UCS is always used in conjunction with a base selector and never optimized in isolation.

\paragraph{Instantiations with existing selectors.}

\begin{table}[t]
\centering
\resizebox{\columnwidth}{!}{
\begin{tabular}{l l}
\toprule
Methods & UCS-augmented scoring rule \\
\midrule
DPP+UCS & $\arg\max_{i\notin S}\ \Delta_{\text{DPP}}(i\mid S)+\lambda(\Phi(S\cup\{i\})-\Phi(S))$ \\
MDL+UCS & $\arg\max_{S\in\mathcal C(x)}\ \text{MDL}(S;x)+\lambda\Phi(S)$ \\
VoteK+UCS & $\text{score}(i)=v(i)+\lambda\log w_{c(i)},\ \ w_c\propto(\widehat g_{n_c}+\epsilon)^{-1}$ \\
\bottomrule
\end{tabular}}
\caption{Instantiations of UCS for three baselines.}
\label{tab:ucs_minimal}
\vspace{-1em}
\end{table}

Table~\ref{tab:ucs_minimal} summarizes how the UCS coverage functional is incorporated into
three standard demonstration selectors.
Across all cases, the UCS functional is added as a lightweight regularization term on top of the original objective, without modifying the underlying retrieval pipeline.
Setting $\lambda=0$ recovers the original formulations. More implementation details are deferred to Appendix \ref{app:ucs_instantiations}.

\section{Experiment Results}
\subsection{Setup}

We evaluate \UCS{} on three intent classification benchmarks: \textsc{BANKING77}~\citep{casanuevaEfficientIntentDetection2020a} (banking-domain intents, $n=10{,}003$), \textsc{CLINC150}~\citep{larsonEvaluationDatasetIntent2019a} (large-scale multi-domain intents, $n=15{,}000$), and \textsc{HWU64}~\citep{LiuESR19} (fine-grained conversational intents, $n=11{,}036$).
To test transfer beyond intent classification, we further evaluate on three Big-Bench Extra Hard (BBEH) reasoning tasks \citep{kazemi2025bbeh}: \emph{Causal Understanding}, \emph{Object Properties}, and \emph{Shuffled Objects}, each containing 200 examples split 80/20 into candidate pool and evaluation set.
For each dataset, demonstrations are selected from the training split, and unless otherwise specified, we fix the selection budget to $B{=}10$ examples per query.

To assess robustness across model scales and families, we consider three frontier LLMs: \texttt{Qwen2.5-7B-Instruct} \citep{Yang2024Qwen25TR}, \texttt{Llama-3.2-3B-Instruct} \citep{dubey2024llama}, and \texttt{Gemma-2-9B-it} \citep{team2024gemma}. We consider three representative demonstration selectors: \textsc{VoteK}, a query-independent baseline, and \textsc{DPP} and \textsc{MDL}, which are query-dependent; \UCS{} is implemented as a plug-in prior on top of each selector, as summarized in Table~\ref{tab:ucs_minimal}, using the OpenICL~\citep{wu2023openicl} codebase\footnote{Our implementation is available at \small{\url{https://github.com/Raina-Xin/UCS/tree/main/OpenICL/openicl/icl_retriever}}}.

\subsection{\UCS{} Improves ICL Demonstration Selection}

\begin{table*}[!t]
\centering
\caption{\textbf{\UCS{} consistently improves base selection across different datasets and models.} ICL accuracy (higher is better) across three intent classification datasets.
Each method selects a demonstration set of size \textbf{budget=10} using \texttt{dict\_dbscan} clustering.
We evaluate on \textbf{test\_size=500} examples and report mean$\pm$std over \textbf{n\_runs=3}.
}
\label{tab:icl_main_multi_dataset}
\resizebox{\textwidth}{!}{%
% We use >{\columncolor{LightCyan}} before the 'c' for the columns we want to highlight.
\begin{tabular}{l|c|c >{\columncolor{LightCyan}}c | c >{\columncolor{LightCyan}}c |c >{\columncolor{LightCyan}}c}
\toprule
\textbf{Backbone Model} & \textbf{Dataset}
& \textbf{MDL} & \textbf{UCS+MDL}
& \textbf{DPP} & \textbf{UCS+DPP}
& \textbf{VoteK} & \textbf{UCS+VoteK} \\
\midrule
\multirow{3}{*}{Qwen2.5-7B-it}
 & Banking77
 & 0.764\scriptsize{$\pm 0.019$} & \textbf{0.771}\scriptsize{$\pm 0.011$}
 & \textbf{0.831}\scriptsize{$\pm 0.005$} & \textbf{0.831}\scriptsize{$\pm 0.007$}
 & 0.518\scriptsize{$\pm 0.008$} & \textbf{0.543}\scriptsize{$\pm 0.020$} \\
 & CLINC150
 & 0.748\scriptsize{$\pm 0.004$} & \textbf{0.752}\scriptsize{$\pm 0.012$}
 & 0.755\scriptsize{$\pm 0.011$} & \textbf{0.775}\scriptsize{$\pm 0.010$}
 & 0.703\scriptsize{$\pm 0.001$} & \textbf{0.744}\scriptsize{$\pm 0.010$} \\
 & HWU64
 & 0.785\scriptsize{$\pm 0.002$} & \textbf{0.801}\scriptsize{$\pm 0.011$}
 & 0.791\scriptsize{$\pm 0.003$} & \textbf{0.794}\scriptsize{$\pm 0.004$}
 & 0.609\scriptsize{$\pm 0.003$} & \textbf{0.671}\scriptsize{$\pm 0.021$} \\
\midrule
\multirow{3}{*}{Llama-3.2-3B-it}
 & Banking77
 & 0.718\scriptsize{$\pm 0.003$} & \textbf{0.732}\scriptsize{$\pm 0.012$}
 & 0.743\scriptsize{$\pm 0.008$} & \textbf{0.746}\scriptsize{$\pm 0.006$}
 & 0.421\scriptsize{$\pm 0.009$} & \textbf{0.426}\scriptsize{$\pm 0.002$} \\
 & CLINC150
 & 0.747\scriptsize{$\pm 0.005$} & \textbf{0.755}\scriptsize{$\pm 0.006$}
 & 0.698\scriptsize{$\pm 0.008$} & \textbf{0.714}\scriptsize{$\pm 0.009$}
 & 0.597\scriptsize{$\pm 0.003$} & \textbf{0.599}\scriptsize{$\pm 0.008$} \\
 & HWU64
 & 0.537\scriptsize{$\pm 0.008$} & \textbf{0.539}\scriptsize{$\pm 0.010$}
 & 0.565\scriptsize{$\pm 0.003$} & \textbf{0.571}\scriptsize{$\pm 0.008$}
 & 0.457\scriptsize{$\pm 0.004$} & \textbf{0.498}\scriptsize{$\pm 0.020$} \\
\midrule
\multirow{3}{*}{Gemma-2-9B-it}
 & Banking77
 & 0.883\scriptsize{$\pm 0.007$} & \textbf{0.893}\scriptsize{$\pm 0.007$}
 & 0.915\scriptsize{$\pm 0.001$} & \textbf{0.916}\scriptsize{$\pm 0.001$}
 & 0.665\scriptsize{$\pm 0.002$} & \textbf{0.677}\scriptsize{$\pm 0.004$} \\
 & CLINC150
 & 0.977\scriptsize{$\pm 0.005$} & \textbf{0.981}\scriptsize{$\pm 0.006$}
 & \textbf{0.990}\scriptsize{$\pm 0.000$} & \textbf{0.990}\scriptsize{$\pm 0.000$}
 & 0.849\scriptsize{$\pm 0.001$} & \textbf{0.852}\scriptsize{$\pm 0.004$} \\
 & HWU64
 & 0.887\scriptsize{$\pm 0.005$} & \textbf{0.892}\scriptsize{$\pm 0.005$}
 & 0.906\scriptsize{$\pm 0.002$} & \textbf{0.907}\scriptsize{$\pm 0.002$}
 & \textbf{0.794}\scriptsize{$\pm 0.002$} & \textbf{0.794}\scriptsize{$\pm 0.002$} \\
\bottomrule
\end{tabular}%
}
\end{table*}

Table~\ref{tab:icl_main_multi_dataset} summarizes ICL accuracy across three intent
classification benchmarks and three backbone models.
Overall, augmenting existing selectors with \UCS{} consistently improves or
matches performance, with the largest gains observed for query-independent
selection.
\ding{182} \textbf{Significant gains for query-independent selection.}
\UCS{} substantially improves \textsc{VoteK}, particularly on challenging tasks.
On \textsc{HWU64}, \UCS{}+VoteK yields accuracy gains of
\textbf{+6.2\%} (Qwen2.5-7B) and \textbf{+4.1\%} (Llama-3.2-3B).
Similarly, on \textsc{CLINC150}, it boosts Qwen2.5-7B performance by \textbf{+4.1\%}.
\ding{183} \textbf{Complementary to query-dependent selectors.}
When combined with \textsc{MDL} or \textsc{DPP}, \UCS{} yields modest but consistent
improvements in several settings (e.g., $\textbf{+1.4\%}$ on
\textsc{Banking77} for Llama-3.2-3B), while preserving performance elsewhere.
\ding{184} \textbf{Robust across models and datasets.}
Across nine dataset--LLM pairs, \UCS{} improves or matches the base selector in
most cases, with larger relative gains in lower-accuracy regimes.
$\triangleright$~\textbf{Takeaway.}
\UCS{} is a lightweight, model-agnostic plug-in that strengthens ICL demonstration
selection by explicitly encouraging coverage of unrevealed clusters.
A detailed runtime breakdown is provided in Appendix~\ref{app:runtime}: offline preprocessing takes 38--57\,s per dataset, and for our recommended pairings (MDL+UCS, VoteK+UCS), the additional online overhead is typically ${\sim}$0--3\,s.

\begin{table*}[!t]
\centering
\caption{\textbf{\UCS{} consistently improves base selection on reasoning tasks (Qwen2.5-7B).} ICL accuracy (higher is better) across three Big-Bench Extra Hard (BBEH) reasoning datasets \citep{kazemi2025bbeh}: Causal Understanding, Object Properties, and Shuffled Objects.
Each method selects a demonstration set of size \textbf{budget=10} using \texttt{dict\_dbscan} clustering.
We evaluate on \textbf{test\_size=40} examples and report mean$\pm$std over \textbf{n\_runs=3}.%Embeddings and evaluation use the corresponding backbone model.
}
\label{tab:icl_main_multi_reasoning}
\resizebox{\textwidth}{!}{%
\begin{tabular}{l|c|c >{\columncolor{LightCyan}}c | c >{\columncolor{LightCyan}}c |c >{\columncolor{LightCyan}}c}
\toprule
\textbf{Backbone Model} & \textbf{Dataset}
& \textbf{MDL} & \textbf{UCS+MDL}
& \textbf{DPP} & \textbf{UCS+DPP}
& \textbf{VoteK} & \textbf{UCS+VoteK} \\
\midrule
\multirow{3}{*}{Qwen2.5-7B-it}
 & Causal
 & 0.608\scriptsize{$\pm 0.012$} & \textbf{0.692}\scriptsize{$\pm 0.042$}
 & 0.517\scriptsize{$\pm 0.012$} & \textbf{0.542}\scriptsize{$\pm 0.012$}
 & 0.492\scriptsize{$\pm 0.012$} & \textbf{0.525}\scriptsize{$\pm 0.00$} \\
 & Properties
 & 0.083\scriptsize{$\pm 0.029$} & \textbf{0.092}\scriptsize{$\pm 0.031$}
 & 0.050\scriptsize{$\pm 0.020$} & \textbf{0.067}\scriptsize{$\pm 0.024$}
 & 0.050\scriptsize{$\pm 0.020$} & \textbf{0.067}\scriptsize{$\pm 0.012$} \\
 & Shuffled
 & 0.200\scriptsize{$\pm 0.043$} & \textbf{0.225}\scriptsize{$\pm 0.000$}
 & 0.133\scriptsize{$\pm 0.012$} & \textbf{0.258}\scriptsize{$\pm 0.012$}
 & 0.242\scriptsize{$\pm 0.063$} & \textbf{0.242}\scriptsize{$\pm 0.043$} \\
% Llama-3.2-3B-it reasoning results omitted (rebuttal)
% Gemma-2-9B-it reasoning results omitted: experiments pending
\bottomrule
\end{tabular}%
}
\end{table*}

Table~\ref{tab:icl_main_multi_reasoning} presents results on three BBEH reasoning tasks.
On Qwen2.5-7B, \UCS{} improves 8 of 9 selector--task settings, with gains up to \textbf{+12.5\,pp} on Shuffled Objects (DPP$\rightarrow$UCS+DPP) and \textbf{+8.4\,pp} on Causal Understanding (MDL$\rightarrow$UCS+MDL), suggesting strong transfer to reasoning tasks.

\subsection{\UCS{} Reveals Cluster ID Distribution Across Tasks and LLMs}

\begin{figure}[htbp]  % You can also use [htbp] instead of [H] if you want more flexible positioning
\centering
\includegraphics[width=0.49\textwidth]{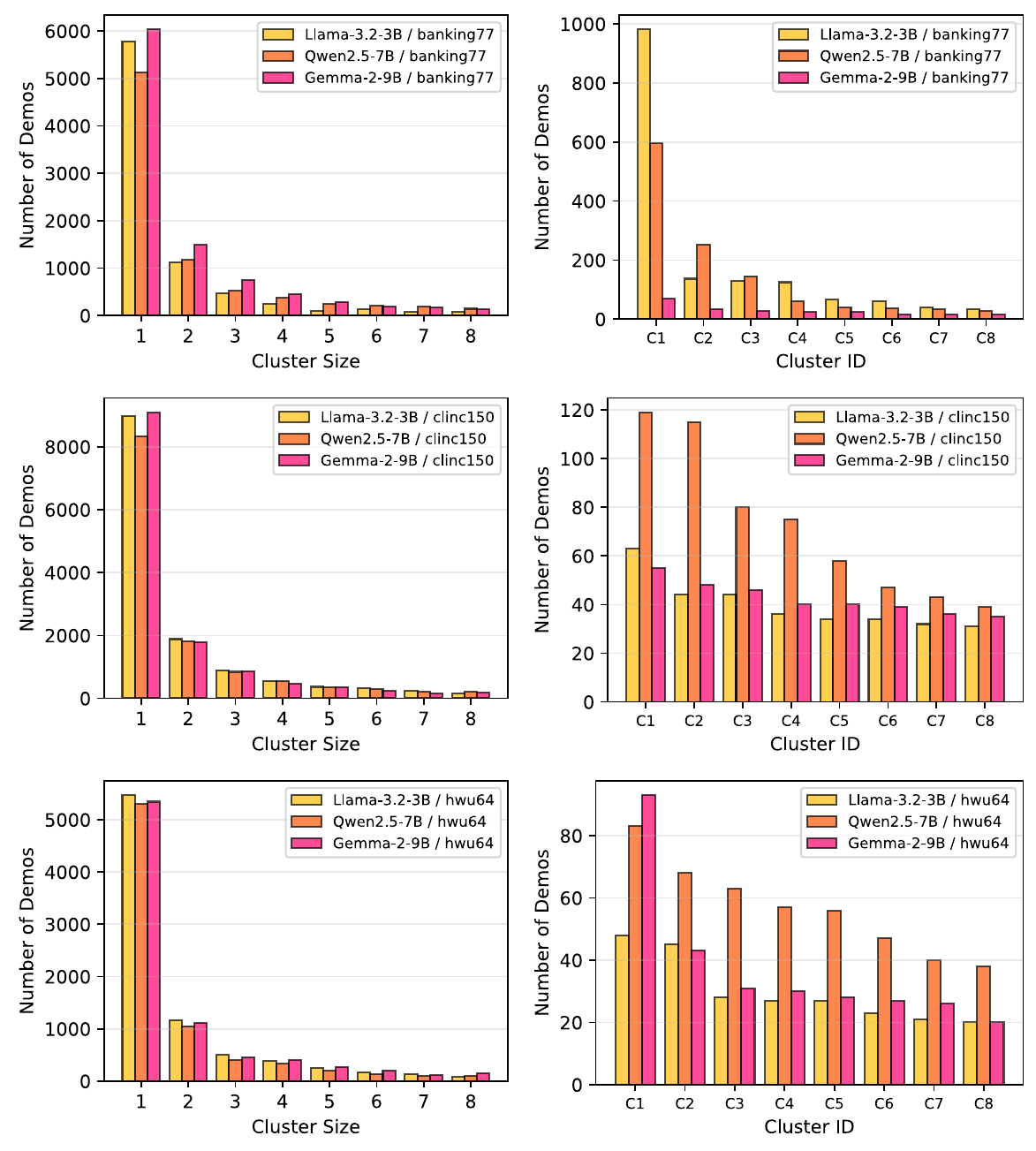}  % Change width and filename as needed
\vspace{-1em}
\caption{\textbf{Cluster-size distributions across datasets and LLMs.}
Left: number of demonstrations in clusters of size $k\in[1,8]$ (heavy-tailed, many singletons).
Right: sizes of the top-$8$ clusters (a few dominant clusters).}
\label{fig:fig3_cluster_comparison_dataset_llms}
\vspace{-1em}
\end{figure}

To characterize how cluster ID is distributed in the demonstration pool, we analyze
the cluster statistics induced by unsupervised clustering.
Figure~\ref{fig:fig3_cluster_comparison_dataset_llms} visualizes the cluster-size
distribution for each dataset--LLM pair, including the mass of small clusters
(left) and the sizes of the largest clusters (right).
Across all settings, the distributions are highly skewed: most demonstrations
belong to small or singleton clusters, while a small number of clusters dominate
the pool.
This long-tailed pattern indicates that a few cluster types are frequently
observed, whereas many others appear rarely.
Importantly, both the degree of skew and the dominant clusters vary substantially
across datasets and LLMs.
$\triangleright$~\textbf{Takeaway.}
Cluster ID distributions vary substantially across tasks and LLMs, motivating
\UCS{} to explicitly account for uneven and model-specific coverage.

\subsection{UCS Prioritizes Cluster Coverage}

To quantify how \UCS{} alters the content of the prompts, we analyze the latent cluster statistics of selected subsets in Table~\ref{tab:coverage_exposure}.
Driven by the SGT-based coverage functional, \UCS{} explicitly biases selection toward the ``rare'' clusters located in the heavy tail of the distribution (as identified in Figure~\ref{fig:fig3_cluster_comparison_dataset_llms}).
This is empirically confirmed by the shift in cluster composition of retrieved demonstrations: \UCS{}-augmented subsets generally exhibit a higher count of unique clusters and a smaller average cluster size compared to their baselines.
This effect is most pronounced for VoteK, where adding \UCS{} eliminates redundancy entirely, reducing the average cluster size from $8.50$ to $1.00$.
By targeting these rare clusters, our method effectively exchanges marginal redundancy for broader exposure to unseen cluster types.
$\triangleright$~\textbf{Takeaway.}
\UCS{} diversifies prompt content by prioritizing the heavy tail of the cluster distribution, ensuring that selected demonstrations maximize coverage of distinct semantic concepts.

\paragraph{Qualitative audit on BANKING77.}
To assess whether the induced clusters are semantically meaningful, we examine representative clusters on \textsc{BANKING77} (Qwen2.5-7B).
Many clusters correspond to coherent, human-recognizable micro-topics: e.g., cluster 380 (size=7) is entirely \emph{unable\_to\_verify\_identity} paraphrases, cluster 298 (size=5) concentrates on \emph{physical card vs.\ virtual card}, and cluster 562 (size=4) is fully \emph{transaction\_charged\_twice}.
When adding UCS on top of DPP, selections shift away from an over-broad dominant cluster (cid=0; size=8{,}970) toward smaller, semantically sharper clusters (e.g., cid=380, identity verification; cid=232, ATM cash issues).
This behavior reflects UCS's frequency-spectrum mechanism: it penalizes over-representation of dominant modes and increases coverage of underrepresented latent clusters.

\begin{table}[h]
\centering
\caption{\textbf{UCS Prioritizes Cluster Coverage.}
Metrics for Qwen on CLINC150. \textbf{\#\,Uniq Clusters}: number of distinct clusters in the selected prompt. \textbf{Cluster Size}: average size of the clusters that the selected demos belong to (over the entire pool). \textbf{1/Size}: per-demo inverse cluster size averaged over the prompt; a value of 1.0 indicates all demos come from singleton clusters.}
\label{tab:coverage_exposure}
\resizebox{\columnwidth}{!}{\begin{tabular}{c | c c c}
\toprule
\textbf{Method} & \textbf{\# Uniq Clusters} & \textbf{Cluster Size} & \textbf{1/Size}  \\
\midrule

MDL & 9.53\scriptsize{$\pm 0.006$} & 6.28\scriptsize{$\pm 0.550$} & 0.689\scriptsize{$\pm 0.007$} \\
UCS+MDL & 9.49\scriptsize{$\pm 0.031$} & 6.57\scriptsize{$\pm 0.734$} & 0.688\scriptsize{$\pm 0.005$} \\
\midrule
DPP & 9.62\scriptsize{$\pm 0.014$} & 6.29\scriptsize{$\pm 0.323$} & 0.701\scriptsize{$\pm 0.003$} \\
UCS+DPP & 9.93\scriptsize{$\pm 0.016$} & 5.39\scriptsize{$\pm 0.178$} & 0.719\scriptsize{$\pm 0.004$} \\
\midrule
VoteK & 9.67\scriptsize{$\pm 0.471$} & 8.50\scriptsize{$\pm 1.390$} & 0.604\scriptsize{$\pm 0.020$} \\
UCS+VoteK & 10.0\scriptsize{$\pm 0.000$} & 1.00\scriptsize{$\pm 0.000$} & 1.00\scriptsize{$\pm 0.000$} \\
\bottomrule
\end{tabular}}
% \vspace{-.8em}
\end{table}

\section{In-depth Analysis of \UCS{}}

\subsection{Joint-Dictionary Extension: Cross-Model Alignment}
\label{subsec:multi_llm_ucs}

As an exploratory extension, we test whether aligning embeddings from multiple LLMs into a shared representation improves robustness.
We extend \UCS{} to a \textbf{Joint Dictionary Learning} framework. Rather than deriving clusters from a single latent space, we align embeddings from multiple LLMs into a shared canonical representation. Formally, given embedding matrices $E^{(m)} \in \mathbb{R}^{N \times d_m}$ from $M$ different models, we learn a shared dictionary $D$ and shared sparse codes $R$ by optimizing orthogonal transformation matrices $B^{(m)}$ for each model. This is achieved by solving a joint reconstruction objective subject to orthogonal constraints:
\begin{equation}
\begin{split}
\min_{D, R, \{B^{(m)}\}} & \sum_{m=1}^M \left\| E^{(m)} B^{(m)} - R D^\top \right\|_F^2, \\
& \text{s.t. } (B^{(m)})^\top B^{(m)} = I.
\end{split}
\label{eq:joint_dict}
\end{equation}
By mapping disparate embedding spaces into this unified coordinate system, we aim to capture semantic invariants agreed upon by multiple models.
We evaluate this approach by constructing UCS priors using a joint dictionary learned from Qwen2.5-7B, Llama-3.2-3B, and Gemma-2-9B. Table~\ref{tab:joint_source_comparison} presents the ICL accuracy comparisons. Contrary to the expectation that multi-model consensus yields more robust priors, the results indicate a performance trade-off. While the joint dictionary maintains parity for the smaller Llama-3.2-3B backbone (e.g., maintaining $74.6\%$ on Banking77 and yielding a marginal gain on HWU64), it causes significant degradation for the stronger Qwen2.5-7B model (e.g., $-11.9\%$ on HWU64). This suggests that forcing disparate embedding manifolds into a shared linear subspace may dilute the fine-grained semantic distinctions present in higher-capacity models, highlighting the difficulty of lossless alignment across heterogeneous architectures.
These results further motivate our default choice of \emph{model-consistent} (single-source) embeddings.
\begin{table}[h]
\centering
\caption{\textbf{Single-Source vs. Joint-Source UCS.} Comparison of \textbf{UCS+DPP} accuracy ($B{=}10$) when using a dictionary derived from the inference model itself (Single-Source) versus a Joint Dictionary aligned across multiple LLMs. The joint representation maintains performance for the 3B model but shows trade-offs for the larger 7B model.}
\label{tab:joint_source_comparison}
\resizebox{\columnwidth}{!}{
\begin{tabular}{c|c|cc|c}
\toprule
\textbf{Backbone} & \textbf{Dataset} & \textbf{Single-Source} & \textbf{Joint-Source} & $\Delta$ \\
% \textbf{Model} & & (UCS+DPP) & (UCS+DPP) & \\
\midrule
\multirow{3}{*}{\shortstack[l]{Qwen}} 
 & Banking77 & 83.1\scriptsize{$\pm 0.7$} & 78.3\scriptsize{$\pm 1.3$} & \textcolor{red}{$-4.8$} \\
 & CLINC150  & 77.5\scriptsize{$\pm 0.1$} & 82.3\scriptsize{$\pm 2.0$} & \textcolor{blue}{$+4.8$} \\
 & HWU64     & 79.4\scriptsize{$\pm 0.4$} & 67.5\scriptsize{$\pm 1.3$} & \textcolor{red}{$-11.9$} \\
\midrule
\multirow{3}{*}{\shortstack[l]{Llama}} 
 & Banking77 & 74.6\scriptsize{$\pm 0.6$} & 74.6\scriptsize{$\pm 1.1$} & $0.0$ \\
 & CLINC150  & 71.4\scriptsize{$\pm 0.9$} & 70.5\scriptsize{$\pm 1.3$} & \textcolor{red}{$-0.9$} \\
 & HWU64     & 57.1\scriptsize{$\pm 0.8$} & 57.7\scriptsize{$\pm 4.7$} & \textcolor{blue}{$+0.6$} \\
% \multirow{3}{*}{\shortstack[l]{Qwen2.5\\7B-Instruct}} 
%  & Banking77 & 0.831\scriptsize{$\pm 0.007$} & 0.783\scriptsize{$\pm 0.013$} & \textcolor{red}{$-0.048$} \\
%  & CLINC150  & 0.871\scriptsize{$\pm 0.008$} & 0.823\scriptsize{$\pm 0.020$} & \textcolor{red}{$-0.048$} \\
%  & HWU64     & 0.794\scriptsize{$\pm 0.004$} & 0.675\scriptsize{$\pm 0.013$} & \textcolor{red}{$-0.119$} \\
% \midrule
% \multirow{3}{*}{\shortstack[l]{Llama-3.2\\3B-Instruct}} 
%  & Banking77 & 0.746\scriptsize{$\pm 0.006$} & 0.746\scriptsize{$\pm 0.011$} & $0.000$ \\
%  & CLINC150  & 0.714\scriptsize{$\pm 0.009$} & 0.705\scriptsize{$\pm 0.013$} & \textcolor{red}{$-0.009$} \\
%  & HWU64     & 0.571\scriptsize{$\pm 0.008$} & \textbf{0.577}\scriptsize{$\pm 0.047$} & \textcolor{blue}{$+0.006$} \\
\bottomrule
\end{tabular}
}
% \vspace{-.8em}
\end{table}

\subsection{Sensitivity Analysis of UCS}
\label{subsec:sensitivity_ucs}

\paragraph{Sensitivity to SGT hyperparameters.}
We examine the robustness of \UCS{} to the hyperparameters of the SGT estimator by varying the extrapolation horizon $t$ and the frequency bin size,
while fixing the base retriever.
As shown in Table~\ref{tab:sgt_sensitivity}, \UCS{} consistently improves over the
base retriever across a wide range of SGT settings.
The paired improvements remain stable as $t$ and the bin size change, indicating
that \UCS{} is robust to different SGT hyperparameters.

\begin{table}[h]
\centering
\caption{\textbf{Sensitivity to SGT hyperparameters.}
Accuracy and improvement for MDL and UCS+MDL for Llama model on Banking77 dataset.}
\label{tab:sgt_sensitivity}
\resizebox{.6\columnwidth}{!}{\begin{tabular}{c c | c c c}
\toprule
\textbf{$ t $} & \textbf{bin size} &
\textbf{$\textsc{UCS}{+}\textsc{MDL}$} &
\textbf{$\Delta$} \\
\midrule
2 & 10  & 73.1\scriptsize{$\pm 1.1$} & \textcolor{blue}{$+$1.3} \\
4 & 10  & 73.0\scriptsize{$\pm 0.7$} & \textcolor{blue}{$+$1.2} \\
8 & 10  & 72.9\scriptsize{$\pm 1.6$} & \textcolor{blue}{$+$1.1} \\
4 & 5   & 72.8\scriptsize{$\pm 1.7$} & \textcolor{blue}{$+$1.0} \\
4 & 20  & 72.9\scriptsize{$\pm 1.2$} & \textcolor{blue}{$+$1.1} \\
\bottomrule
\end{tabular}}
% \vspace{-.8em}
\end{table}

\paragraph{Sensitivity to selection budget.}
We vary the number of in-context demonstrations to examine whether \UCS{} remains
effective across different selection budget.
As shown in Table~\ref{tab:budget_sensitivity}, \UCS{} consistently improves the base retriever across a wide range of budgets.
The gains are more pronounced at moderate to larger budgets, where selection
quality plays a greater role, while performance remains stable in the low-budget
regime.
These results indicate that \UCS{} provides robust performance benefits that could generalize across different demonstration selection budget.

\begin{table}[!ht]
\centering
\caption{\textbf{Sensitivity to selection budget.}
Accuracy (\%) as a function of the number of demonstrations for MDL and UCS+MDL with Llama on Banking77 dataset.}
\label{tab:budget_sensitivity}
\resizebox{.7\columnwidth}{!}{\begin{tabular}{c | c c c}
\toprule
\textbf{Budget} &
\textbf{$\textsc{MDL}$} &
\textbf{$\textsc{UCS}{+}\textsc{MDL}$ }&
\textbf{$\Delta$} \\
\midrule
0  & 34.7\scriptsize{$\pm 0.7$} & 34.9\scriptsize{$\pm 0.2$} & \textcolor{blue}{$+$0.2} \\
1  & 56.4\scriptsize{$\pm 1.1$} & 56.5\scriptsize{$\pm 1.6$} & \textcolor{blue}{$+$0.1} \\
5  & 70.9\scriptsize{$\pm 1.9$} & 72.4\scriptsize{$\pm 1.1$} & \textcolor{blue}{$+$1.5} \\
10 & 71.8\scriptsize{$\pm 0.3$} & 73.2\scriptsize{$\pm 1.2$} & \textcolor{blue}{$+$1.4} \\
20 & 63.6\scriptsize{$\pm 0.3$} & 65.9\scriptsize{$\pm 2.1$} & \textcolor{blue}{$+$2.3} \\
\bottomrule
\end{tabular}}
\vspace{-.1em}
\end{table}

\section{Ablation Studies}
\label{sec:ablation}

To isolate the impact of core design choices in \UCS{}, we conduct controlled ablations on embedding construction, cluster definition, and scoring design using the Llama model (Table~\ref{tab:ablation_ucs}).
\paragraph{Embedding Construction.}
Results indicate that \UCS{} is robust to embedding extraction choices, though last-layer representations with mean pooling consistently yield the best results.
Mid-layer features and last-token pooling degrade performance, suggesting that aggregated, final-layer semantics best capture latent task structure.
Interestingly, applying $\ell_2$ normalization (\textit{Last+Mean+$\ell_2$Norm}) offers slight gains over the unnormalized default, indicating that directional consistency aids density-based clustering.
\paragraph{Cluster Definition.}
We compare our dictionary-based clustering against direct embedding clustering without dictionary learning (\textit{dbscan}) and simple atom assignment (\textit{dict\_argmax}).
\UCS{} outperforms both baselines, particularly on HWU64 ($+2.9\%$ vs.\ \textit{dbscan}).
This confirms that projecting into a sparse dictionary space effectively denoises the representation, while density-based clustering preserves local semantic structures that simple argmax assignment destroys.
\paragraph{Importance of SGT Scoring.}
Replacing the full UCS objective with coverage-only ($n_{\text{seen}}$) leads to significant performance drops (e.g., $-2.8\%$ on HWU64).
The superior performance of the combined default objective confirms that the SGT-derived unseen estimator is non-redundant: it guides selection toward both seen and unseen coverage that simple coverage metrics fail to anticipate.

\begin{table}[!t]
\centering
\caption{\textbf{Ablation results}: Accuracy in \%, mean$\pm$std over 3 runs on three datasets. ``Mid'' and ``Last'' refer to the layer used for embedding extraction. Two alternative clustering methods are described in Section \ref{sec:ucs_units}.}
\label{tab:ablation_ucs}
\resizebox{\columnwidth}{!}{%
\begin{tabular}{l|ccc}
\toprule
\textbf{\UCS{}+DPP} & \textbf{Banking77} & \textbf{CLINC150} & \textbf{HWU64} \\
\midrule
\multicolumn{4}{c}{\textit{Embedding construction}} \\
Mid+Mean                 & 73.9\scriptsize{$\pm$0.1} & 70.7\scriptsize{$\pm$0.8} & 57.3\scriptsize{$\pm$1.2} \\
Last+LastToken           & 73.7\scriptsize{$\pm$1.4} & 71.7\scriptsize{$\pm$0.5} & 56.7\scriptsize{$\pm$1.5} \\
Last+Mean+$\ell_2$Norm   & 74.8\scriptsize{$\pm$1.3} & 71.7\scriptsize{$\pm$0.9} & 57.5\scriptsize{$\pm$0.8} \\
\midrule
\multicolumn{4}{c}{\textit{Clustering method}} \\
dbscan                   & 73.6\scriptsize{$\pm$1.1} & 70.4\scriptsize{$\pm$0.2} & 54.2\scriptsize{$\pm$0.6} \\
dict\_argmax             & 74.0\scriptsize{$\pm$0.7} & 69.6\scriptsize{$\pm$0.1} & 55.8\scriptsize{$\pm$1.1} \\
\midrule
\multicolumn{4}{c}{\textit{UCS score design}} \\
$n_{\text{seen}}$                     & 74.6\scriptsize{$\pm$0.1} & 70.2\scriptsize{$\pm$0.7} & 54.3\scriptsize{$\pm$0.3} \\
$\hat n_{\text{unseen}}$              & 74.4\scriptsize{$\pm$0.6} & 70.3\scriptsize{$\pm$0.7} & 54.1\scriptsize{$\pm$0.8} \\
$\hat n_{\text{unseen}}/n_{\text{seen}}$ & 74.7\scriptsize{$\pm$0.5} & 70.1\scriptsize{$\pm$0.6} & 54.0\scriptsize{$\pm$0.8} \\
\midrule
UCS                 & 74.6\scriptsize{$\pm$0.6} & 71.4\scriptsize{$\pm$0.9} & 57.1\scriptsize{$\pm$0.8} \\
\bottomrule
\end{tabular}}
\end{table}

\paragraph{D. Rarity-Only VoteK Controls.}
To test whether UCS+VoteK gains are simply due to rarity weighting, we evaluate two rarity-only controls that keep the VoteK protocol but \emph{remove} the Good--Turing/SGT extrapolation:
\textbf{B1} (inverse cluster-size rarity) adds a bonus depending only on cluster size, and
\textbf{B2} (spectrum rarity without GT) uses the size spectrum but explicitly drops the GT ratio/discounting.
Table~\ref{tab:rarity_controls} shows that neither control matches UCS+VoteK: the rarity-only baselines are inconsistent across datasets and fail to reproduce UCS+VoteK improvements, supporting that gains come from the SGT-based extrapolation rather than naive rarity weighting.

\begin{table}[!t]
\centering
\caption{\textbf{Rarity-only VoteK controls} (Llama-3.2-3B-it). Mean$\pm$std accuracy (\%) over 3 runs. B1: inverse-size rarity. B2: spectrum rarity without GT.}
\label{tab:rarity_controls}
\resizebox{\columnwidth}{!}{%
\begin{tabular}{l|ccc}
\toprule
\textbf{Method} & \textbf{Banking77} & \textbf{CLINC150} & \textbf{HWU64} \\
\midrule
VoteK              & 42.1\scriptsize{$\pm$0.9} & 59.7\scriptsize{$\pm$0.3} & 45.7\scriptsize{$\pm$0.4} \\
UCS+VoteK          & 42.6\scriptsize{$\pm$0.2} & 59.9\scriptsize{$\pm$0.8} & 49.8\scriptsize{$\pm$2.0} \\
B1 & 40.8\scriptsize{$\pm$2.3} & 60.0\scriptsize{$\pm$1.4} & 44.2\scriptsize{$\pm$1.1} \\
B2 & 43.6\scriptsize{$\pm$1.7} & 59.1\scriptsize{$\pm$1.1} & 44.2\scriptsize{$\pm$1.1} \\
\bottomrule
\end{tabular}}
\vspace{-.8em}
\end{table}

\section{Conclusion}
We proposed \UCS{}, a statistically grounded framework for improving in-context
demonstration selection via a subset-level \emph{coverage} prior.
\UCS{} discretizes LLM embeddings into latent clusters and uses a Smoothed
Good--Turing estimator to quantify how much seen and unrevealed clusters a candidate subset covers under the same candidate-pool sampling process.
Across three intent classification benchmarks, three BBEH reasoning tasks, and three frontier LLMs, \UCS{}
consistently improves or matches strong baselines, with the largest gains for
query-independent selection %(up to \textbf{$+6.2\%$} when augmenting \textsc{VoteK})
and stable improvements for query-dependent selectors.
Beyond accuracy gains, \UCS{} offers interpretable insights into cluster
distributions across tasks and models, highlighting the long-tailed and
model-specific nature of latent cluster distribution.
Overall, \UCS{} provides a lightweight, training-free plug-in for more reliable
ICL selection by explicitly encouraging broader latent cluster coverage.
\section*{Limitations}
%Limitations section is mandatory for ACL and it is not counted into the page limit
First, \UCS{} discretizes continuous LLM embeddings into latent clusters
using unsupervised methods.
While this design is flexible and training-free, the resulting units may not
always align with human-interpretable concepts.
Future work could explore alternative or hybrid unit discovery strategies,
including supervised signals, semantic grounding, or multi-granular
representations.

Second, while we have extended evaluation to three BBEH reasoning tasks, the
majority of our experiments focus on intent classification with relatively
short demonstrations and contexts.
Extending \UCS{} to tasks with longer inputs, generation-heavy objectives, or structured outputs
(e.g., mathematical reasoning) remains a promising direction.

Third, \UCS{} is designed to be model-consistent and therefore requires embedding
the demonstration pool using the LLM at evaluation time.
This incurs additional offline computation for large pools, but also opens up
opportunities for future work on shared or aligned representations across
multiple models.

Finally, in this work we only combine \UCS{} with widely used and well-established
baseline selectors (e.g., VoteK, DPP, MDL) to provide a clear proof of concept.
Recent reinforcement-learning-based selection methods offer
additional opportunities for integration.
Exploring how \UCS{} can be incorporated into policy-based iterative selection
pipelines—such as by serving as a coverage-aware reward or regularizer—is an
exciting avenue for future research.

 % limiations is mandatory for ACL

\bibliography{custom_CR}
\appendix
\newpage
\section{Additional Details for the UCS Algorithm}
\label{app:ucs_details}

\subsection{Embedding Details}
\label{app:embedding_details}

\noindent \textbf{Pooling choice.}
In the main paper we use masked mean pooling (Eq.~\ref{eq:masked-mean-pooling}) to obtain a single vector per input.
We also experimented with alternative pooling rules such as taking the first token embedding or the last non-padding token embedding;
these variants produced similar qualitative trends but were slightly less stable across models.

\noindent \textbf{Layer selection.}
Unless otherwise specified, we extract hidden states from a fixed layer $\ell$ of the same LLM used for ICL evaluation.
In practice, using either the final layer or a mid-layer works well; we fix $\ell$ across all methods for a fair comparison.

\noindent \textbf{Normalization.}
We do not apply $\ell_2$ normalization in our main experiments.
(When enabled, normalization can reduce scale effects for cosine-distance-based clustering, but it was unnecessary in our setup.)

\noindent \textbf{Optional PCA.}
For speed and memory, we optionally apply PCA to reduce embedding dimensionality before downstream clustering/dictionary learning.
When enabled, PCA is fit on the training pool embeddings only and applied to the same pool; this does not use labels.

\noindent \textbf{Batching and throughput notes.}
Embedding is computed in mini-batches with truncation to a maximum sequence length.
When the embedding model is sharded across multiple devices (e.g., \texttt{device\_map=auto}), we place input tensors on the device that
hosts the token embedding layer to avoid unnecessary transfers.
We also disable KV-cache during embedding extraction (since generation is not needed), which improves stability for some architectures.

\subsection{DBSCAN Radius Selection via $k$NN Quantiles}
\label{app:dbscan_eps}

DBSCAN requires a neighborhood radius $\texttt{eps}$ and a minimum number of neighbors $\texttt{min\_samples}$.
We set $\texttt{eps}$ \emph{data-adaptively} using a simple $k$NN quantile heuristic that is robust across datasets and embedding scales.

\paragraph{Cosine distance in normalized code space.}
We run DBSCAN on the normalized dictionary codes $\{\bar r_i\}_{i=1}^N$ (Eq.~\ref{eq:code_norm}) using cosine distance
\[
d_{\cos}(\bar r_i,\bar r_j) \;=\; 1 - \frac{\langle \bar r_i,\bar r_j\rangle}{\|\bar r_i\|_2\|\bar r_j\|_2}.
\]
Since $\bar r_i$ is already $\ell_2$-normalized, cosine distance reduces to $1-\langle \bar r_i,\bar r_j\rangle$.

\paragraph{Heuristic for choosing $\texttt{eps}$.}
For each point $i$, we compute its $k$-th nearest-neighbor distance under cosine distance:
\[
d_i^{(k)} \;=\; \text{$k$-NN distance of }\bar r_i \text{ among } \{\bar r_j\}_{j\neq i}.
\]
We then set
\begin{equation}
\texttt{eps} \;=\; \mathrm{Quantile}_q\Big(\{d_i^{(k)}\}_{i=1}^N\Big),
\label{eq:eps_quantile}
\end{equation}
where $q\in(0,1)$ is a user-controlled quantile (e.g., $q\in[0.05,0.15]$ in our sweeps) and $k$ is a small constant
(e.g., $k=10$).
Intuitively, $d_i^{(k)}$ estimates a local neighborhood radius around point $i$; taking a low quantile yields an $\texttt{eps}$
that is tight enough to avoid merging unrelated patterns, while still admitting dense ``head'' regions of the code manifold.
This heuristic automatically adapts to the intrinsic spread of the codes, removing the need for dataset-specific
distance calibration.

\paragraph{Practical knobs.}
In code space, we use a slightly larger $\texttt{min\_samples}$ than in raw embedding DBSCAN to discourage spurious micro-clusters
formed by isolated points (e.g., $\texttt{min\_samples}\ge 5$).
When DBSCAN returns noise points ($c_i=-1$), we map each of them to a new singleton, ensuring every example contributes
a valid discrete type count for downstream SGT estimation.

\subsection{Why Dictionary+DBSCAN Rather than Argmax Atom Assignment}
\label{app:dict_vs_argmax}

A natural discretization after dictionary learning is to assign each example to its largest-magnitude atom,
\[
c_i \;=\; \arg\max_{j\in[K]} |r_{ij}|.
\]
This \emph{argmax-atoms} scheme is fast and yields exactly $K$ types, but we found it to be a weaker inductive bias
for inducing clusters than clustering in code space.

\paragraph{(1) Argmax collapses compositional structure.}
Dictionary codes are not meant to be purely one-hot: many embeddings are better described by \emph{combinations} of atoms.
Two examples may share the same dominant atom but differ substantially in the remaining activated atoms; argmax forces them
into the same unit even if their latent patterns differ.
DBSCAN on the full (normalized) code vectors can separate such cases by grouping on \emph{overall code similarity} rather than
a single coordinate.

\paragraph{(2) Heavy-tailed atom usage distorts type counts.}
In practice, atom activations exhibit a heavy-tailed frequency spectrum: a few atoms become ``head'' atoms used by many points,
while many atoms are rarely used.
Argmax assignment tends to over-concentrate mass on head atoms, producing a small number of very large clusters and a long tail
of tiny clusters. This concentration reduces the effective number of types observed in a subset, which can make unseen estimation
less informative under a fixed budget.
Clustering in code space mitigates this by allowing multiple distinct clusters to exist within the region dominated by a head atom,
as long as the full code patterns differ.

\paragraph{(3) Better alignment with the ``type'' abstraction needed by SGT.}
SGT-style estimators treat each unit as an abstract type that can be observed multiple times in a sample.
A good type system should be (i) stable to small perturbations, (ii) fine-grained enough to distinguish qualitatively different
patterns, and (iii) coarse enough that repeated observations occur at realistic subset sizes.
Argmax is often too coarse in high-density regions (merging distinct patterns) and too brittle around ties
(small coefficient changes can flip the argmax atom).
DBSCAN on normalized codes provides a more stable middle ground: it forms repeatable head clusters where patterns recur,
and allocates ambiguous or idiosyncratic points to tail clusters or the noise unit.

\paragraph{Connection to our ridge codes.}
We use ridge (L2) coding (Eq.~\ref{eq:ridge_code}) specifically to avoid overly sparse, near-one-hot codes that make argmax
assignment almost deterministic and distance computations brittle.
With smoother ridge codes, cosine distances in code space become meaningful, which improves DBSCAN’s ability to discover
consistent latent units.

\paragraph{Takeaway.}
Argmax-atoms is a useful baseline discretization, but dictionary+DBSCAN better respects compositional cluster signatures
and yields a unit system that empirically supports stronger coverage estimates and downstream selection.

\subsection{Smoothed Good--Turing Estimator Details}
\label{app:ucs_estimator_details}

This appendix specifies the stabilized Smoothed Good--Turing (SGT) procedure used to compute
$\widehat{U}_t(S)$ from the subset spectrum $\{f_s(S)\}_{s\ge1}$.
We follow the classical Good--Turing estimator but incorporate standard finite-sample stabilizations:
(i) truncation/binning of high counts, (ii) an offset to avoid unstable extrapolation, and
(iii) optional smoothing of the frequency spectrum.

\paragraph{Inputs.}
Given a candidate subset $S$, we compute the multiplicities $n_u(S)$ and the spectrum
$f_s(S)=|\{u:n_u(S)=s\}|$ for $s\ge1$.
We optionally exclude a designated noise label (e.g., DBSCAN noise) from the type universe
before forming $\{f_s(S)\}$.

\paragraph{Classical Good--Turing functional.}
Let $m>0$ denote the number of additional samples we wish to extrapolate, and define the expansion factor $t \coloneqq m/|S|$ (e.g., $t=5$ means extrapolating to $5\times|S|$ additional draws).
The (unsmoothed) Good--Turing estimator for the number of unseen types after $m$ additional draws is
\begin{equation}
\widehat{U}^{\mathrm{GT}}_{t}(S)
=
-\sum_{s\ge1} (-t)^s\, f_s(S).
\label{eq:gt_raw}
\end{equation}
In finite samples, the alternating series in~\eqref{eq:gt_raw} can have high variance for moderate/large $t$,
motivating stabilized variants.

\paragraph{Truncation / binning.}
To reduce sensitivity to large multiplicities (which are typically sparse), we truncate the spectrum at a
maximum count $M$ (denoted \texttt{bin\_size} in code):
\begin{equation}
\widehat{U}^{\mathrm{GT}}_{t,M}(S)
=
-\sum_{s=1}^{M} (-t)^s\, f_s(S),
\label{eq:gt_trunc_app}
\end{equation}
treating $f_s(S)=0$ for $s>M$.
We use a small default $M$ (e.g., 20) throughout experiments, and report sensitivity in ablations.

\paragraph{Offset-stabilized extrapolation.}
Following standard stabilization heuristics, we introduce an \emph{offset} parameter $\alpha\in[1,2]$
(denoted \texttt{offset} in code) that damps the contribution of higher-order terms for extrapolation.
Concretely, we reweight each term in~\eqref{eq:gt_trunc_app} by a tail probability factor
$w_s(t,\alpha)\in[0,1]$ that decreases with $s$:
\begin{equation}
\widehat{U}^{\mathrm{SGT}}_{t,M}(S)
=
-\sum_{s=1}^{M} (-t)^s\, w_s(t,\alpha)\, f_s(S).
\label{eq:sgt_weighted_app}
\end{equation}
Intuitively, $w_s(t,\alpha)$ acts as a soft truncation that discounts unstable high-order terms when
extrapolating beyond the observed sample.
In our implementation, $w_s(t,\alpha)$ is computed via a simple binomial tail expression
(as in \texttt{utils\_sgt.py}), and we fix $\alpha$ to a default value (e.g., $\alpha=1.0$) unless stated otherwise.

\paragraph{Optional spectrum smoothing.}
The raw spectrum $\{f_s(S)\}$ can be noisy, especially for small subsets.
Optionally, we smooth the nonzero portion of the spectrum by fitting a simple parametric model
(e.g., power-law) and replacing $f_s(S)$ by a smoothed $\widetilde{f}_s(S)$ before applying~\eqref{eq:sgt_weighted_app}.
When enabled, we fit the model to the first $K$ bins (default $K=M$) and clamp negative fitted values to zero:
\begin{equation}
f_s(S)\;\leftarrow\;\widetilde{f}_s(S), \qquad s=1,\dots,M.
\label{eq:spectrum_smooth}
\end{equation}
This option is primarily used as a robustness knob; our main experiments use smoothing only when explicitly noted.

\paragraph{Non-negativity and numerical guards.}
Since $\widehat{U}_t(S)$ represents an unseen \emph{count}, we enforce non-negativity:
\begin{equation}
\widehat{U}_t(S)
=
\max\!\big(0,\; \widehat{U}^{\mathrm{SGT}}_{t,M}(S)\big).
\label{eq:clamp}
\end{equation}
If the estimate is non-finite (NaN/Inf) due to numerical issues, we also set it to zero.
These guards match our selection-time implementation, ensuring that the UCS score
$\widehat{K}_t(S)=K_{\mathrm{seen}}(S)+\widehat{U}_t(S)$ remains well-defined for all subsets.

\paragraph{Hyperparameters.}
We summarize the estimator hyperparameters used throughout the paper:
\begin{itemize}
    \item Expansion factor: $t$ (default $t=5$).
    \item Truncation / bin size: $M$ (default $M=20$).
    \item Offset: $\alpha\in[1,2]$ (default $\alpha=1.0$).
    \item Optional spectrum smoothing: on/off (default off unless stated).
\end{itemize}
We either fix these values across datasets or tune lightly on a small validation split;
details are provided alongside experimental settings.

\section{Runtime Breakdown}
\label{app:runtime}

Table~\ref{tab:runtime_offline} reports the one-time offline preprocessing cost (embedding + dictionary learning + DBSCAN), and Table~\ref{tab:runtime_online} reports the end-to-end online overhead (selection + ICL evaluation) when adding UCS.
All experiments use a single NVIDIA A100 80GB GPU.
For our recommended pairings (MDL+UCS and VoteK+UCS), the additional end-to-end time is typically ${\sim}$0--3\,s.
DPP+UCS incurs larger overhead (+600--680\,s) due to recomputation inside the greedy loop; we treat MDL/VoteK+UCS as the default efficient instantiations.

\begin{table}[t]
\centering
\caption{\textbf{Offline preprocessing} (one-time per dataset+model, in seconds).}
\label{tab:runtime_offline}
\resizebox{\columnwidth}{!}{%
\begin{tabular}{l|cccc|c}
\toprule
\textbf{Dataset (size)} & \textbf{Pool embed} & \textbf{Dict fit} & \textbf{Dict transform} & \textbf{DBSCAN} & \textbf{Total} \\
\midrule
CLINC150 (${\sim}$15k) & 35.88 & 19.65 & 0.17 & 0.37 & 57.03 \\
HWU64 (${\sim}$10k)    & 19.87 & 16.57 & 0.24 & 0.24 & 37.83 \\
BANKING77 (${\sim}$11k) & 27.70 & 14.25 & 0.19 & 0.26 & 43.31 \\
\bottomrule
\end{tabular}}
\end{table}

\begin{table}[t]
\centering
\caption{\textbf{Online selection overhead} ($\Delta$ end-to-end time in seconds, selection + ICL evaluation).}
\label{tab:runtime_online}
\resizebox{\columnwidth}{!}{%
\begin{tabular}{l|ccc}
\toprule
\textbf{Dataset} & \textbf{MDL+UCS} & \textbf{VoteK+UCS} & \textbf{DPP+UCS} \\
\midrule
CLINC150  & +1.96  & $-$13.37 & +605.84 \\
HWU64     & $-$0.09 & +2.75   & +680.70 \\
BANKING77 & +2.28  & +1.19   & +625.38 \\
\bottomrule
\end{tabular}}
\end{table}

\section{Details of UCS Instantiations}
\label{app:ucs_instantiations}
\subsection{Cluster assignments and frequency-of-frequencies}
\label{app:ucs_prelim}

Each training example $i$ is assigned to a discrete cluster
$c(i)\in\mathcal C$ as described in Section~\ref{sec:ucs_units}.
A designated noise label (e.g., DBSCAN noise) is excluded from all UCS computations below.

We use two types of frequency summaries.

\paragraph{Subset-level frequency-of-frequencies (DPP and MDL).}
For a candidate subset $S$, let
\begin{equation}
n_u(S)
=
\big|\{\, i\in S : c(i)=u \,\}\big|
\end{equation}
denote the multiplicity of cluster $u$ in $S$.
The corresponding frequency-of-frequencies (FoF) is defined as
\begin{equation}
f_s(S)
=
\big|\{\, u : n_u(S)=s \,\}\big|.
\end{equation}

\paragraph{Corpus-level cluster-size spectrum (VoteK).}
Over the entire training pool, let
\begin{equation}
n_u = \big|\{\, i : c(i)=u \,\}\big|
\end{equation}
denote the global size of cluster $u$.
We define the corpus-level frequency-of-frequencies (FoF) as
\begin{equation}
g_s = \big|\{\, u : n_u=s \,\}\big|.
\end{equation}
The spectrum $\{g_s\}$ characterizes the global distribution of cluster sizes in the training corpus.

% The subset-level spectrum $\{f_s(S)\}$ characterizes the internal composition of a
% candidate demonstration set, while the corpus-level FoF $\{g_s\}$ summarizes the
% global distribution of cluster sizes in the training corpus.

\subsection{Subset-level UCS for DPP and MDL}
\label{app:ucs_subset}

For query-dependent selectors (DPP and MDL), UCS operates at the subset level.

\paragraph{UCS coverage functional.}
Given a subset $S$, we define the UCS coverage score as
\begin{equation}
\Phi(S)
=
K_{\text{seen}}(S)
+
\widehat U_t(S),
\label{eq:ucs_phi}
\end{equation}
where
\begin{equation}
K_{\text{seen}}(S)
=
\big|\{\, u : n_u(S)>0 \,\}\big|
\end{equation}
is the number of distinct clusters observed in $S$.
The term $\widehat U_t(S)$ estimates the number of previously unseen clusters that would
be revealed after observing $t>0$ additional samples drawn from the same underlying
process as $S$.
This estimate is computed from the subset-level frequency-of-frequencies
$\{f_s(S)\}_{s\ge1}$ using a stabilized Smoothed Good--Turing estimator.
Negative or non-finite estimates are clamped to zero for numerical stability.

Intuitively, $\Phi(S)$ favors subsets that both cover many distinct clusters and exhibit
frequency patterns indicative of additional latent coverage.

\paragraph{DPP + UCS.}
For DPP, demonstrations are selected greedily.
At each step, we select the example $i\notin S$ that maximizes
\begin{equation}
\Delta_{\text{DPP}}(i\mid S)
+
\lambda\big(\Phi(S\cup\{i\})-\Phi(S)\big),
\end{equation}
where $\Delta_{\text{DPP}}(i\mid S)$ is the standard DPP marginal gain.
The UCS term therefore rewards candidates that substantially increase the estimated
coverage of the growing subset.

\paragraph{MDL + UCS.}
For MDL, we score complete candidate subsets rather than individual additions.
Each subset $S$ is assigned the score
\begin{equation}
\text{MDL}(S;x)
+
\lambda\,\Phi(S),
\end{equation}
where $\text{MDL}(S;x)$ is the original MDL objective based on label uncertainty.
UCS thus acts as a subset-level regularizer, encouraging MDL to prefer subsets with
broader estimated coverage.
Candidate subsets are generated using the same proposal mechanism as the original
MDL retriever.

\subsection{Corpus-level UCS prior for VoteK}
\label{app:ucs_votek}

VoteK selects a single, query-independent demonstration set. Accordingly, UCS is instantiated as a factorized corpus-level rarity prior derived via Smoothed Good-Turing (SGT) estimation.

\paragraph{SGT Prior Computation.}
We first compute the corpus-level FoF spectrum $\{g_s\}_{s\ge1}$. To estimate the probability mass of clusters with size $s$, we apply the SGT estimator:

1. \textbf{Smoothing:} We fit a parametric model (e.g., Power-law, Poisson, or Negative Binomial) to $\{g_s\}$ to obtain a non-negative smoothed spectrum $\{\widehat g_s\}_{s\ge1}$.

2. \textbf{Good-Turing Adjustment:} We compute the adjusted count $s^*$ for size $s$ using the Good-Turing rule:
\begin{equation}
s^* = (s+1) \frac{\widehat g_{s+1}}{\widehat g_s}.
\end{equation}

3. \textbf{Weight Assignment:} The estimated probability mass for a cluster of size $s$ is given by $p(s) = s^*/N$, where $N$ is the total number of examples. Each cluster $u$ is assigned a weight inversely proportional to this mass:
\begin{equation}
w_u \propto \frac{1}{p(n_u) + \varepsilon},
\label{eq:cluster_weight}
\end{equation}
where $n_u$ is the size of cluster $u$ and $\varepsilon$ ensures stability. Weights are normalized such that their mean over non-ignored clusters is one.

\paragraph{VoteK + UCS scoring.}
Given the standard VoteK vote count $v(i)$ for example $i$, we define the final score as
\begin{equation}
\text{score}(i) = v(i) + \lambda\log w_{c(i)}.
\end{equation}
When $\lambda=0$, this recovers the original VoteK ranking. For $\lambda>0$, the additional term promotes examples from clusters with globally rare sizes (as estimated by SGT), enhancing the diversity of the selected corpus-level demonstration set.

\section{Hyperparameter Settings for Reproducibility}

For reproducibility, we report the complete set of hyperparameters used in our experiments.
Table~\ref{tab:hyperparams-common} summarizes the common settings shared across all experiments, including budget, batch sizes, and dictionary learning configurations.

While most parameters are fixed, we tune two key hyperparameters using grid search to adapt to the semantic density of different embedding spaces and datasets:
\begin{itemize}
    \item \texttt{dbscan\_q}: The density quantile threshold for the dictionary-based DBSCAN clustering. This determines the granularity of the discovered clusters. Tuned values are reported in Table~\ref{tab:params_dbscan_q}.
    \item \texttt{sgt\_lambda} ($\lambda$): The weight of the UCS unseen-coverage term added to the Retriever's score. This balances the trade-off between the base objective (e.g., diversity in DPP, compression in MDL) and the exploration of rare/unseen clusters. Tuned values for each retriever-model-dataset combination are listed in Table~\ref{tab:params_sgt_lambda}.
\end{itemize}

Unless explicitly varied in ablation studies, all other settings follow Table~\ref{tab:hyperparams-common}.

\begin{table*}[t]
\centering
\caption{\textbf{Common Hyperparameters used for Experiments.}}
\label{tab:hyperparams-common}
\resizebox{\textwidth}{!}{%
\begin{tabular}{l l l}
\toprule
\textbf{Component} & \textbf{Hyperparameter } & \textbf{Value} \\
\midrule

\multirow{6}{*}{Data / Evaluation}
& \texttt{--dataset\_type} & banking77 / clinc150 / hwu64 \\
& \texttt{--budget} & 10 \\
& \texttt{--test\_size} & 500 \\
& \texttt{--n\_runs} & 3 \\
& \texttt{--seed} & 42 \\
& \texttt{--icl\_batch\_size} & 16 \\
\midrule

\multirow{4}{*}{Backbone / Embeddings}
& \texttt{--endpoint\_name} & Qwen/Qwen2.5-7B-Instruct, meta-llama/Llama-3.2-3B-Instruct, google/gemma-2-9b-it \\
& \texttt{--embedding\_model\_name} & Qwen/Qwen2.5-7B-Instruct, meta-llama/Llama-3.2-3B-Instruct, google/gemma-2-9b-it \\
& \texttt{--sentence\_transformers\_model} & all-mpnet-base-v2 \\
& \texttt{--candidate\_num} & 50 \\
\midrule

\multirow{9}{*}{Dictionary Learning}
& \texttt{--dict\_n\_components} & 64 \\
& \texttt{--dict\_alpha} & 10.0 \\
& \texttt{--dict\_top\_k} & 4 \\
& \texttt{--dict\_tau} & $10^{-3}$ \\
& \texttt{--dict\_transform\_algorithm} & omp \\
& \texttt{--dict\_regularization\_type} & l2 \\
& \texttt{--dict\_max\_iter} & 50 \\
& \texttt{--dict\_pca\_dim} & 128 \\
& \texttt{--dict\_batch\_size} & 512 \\
\midrule

\multirow{5}{*}{Clustering (dict\_dbscan)}
& \texttt{--clustering} & dict\_dbscan \\
& \texttt{--dbscan\_k} & 20 \\
& \texttt{--dbscan\_q} & 0.01 \\
& \texttt{--dbscan\_min\_samples} & 1 \\
\midrule

\multirow{4}{*}{SGT / UCS}
& \texttt{--sgt\_lambda} & 0.1 \\
& \texttt{--sgt\_t} & 5 \\
& \texttt{--sgt\_bin\_size} & 20 \\
& \texttt{--sgt\_offset} & 1.0 \\
\midrule

\multirow{5}{*}{Base Selectors}
& \texttt{--baselines} & mdl / mdl\_sgt / dpp / dpp\_sgt / votek / votek\_sgt \\
& \texttt{--votek\_k} & 3 \\
& \texttt{--mdl\_select\_time} & 5 \\
& \texttt{--mdl\_ce\_model} & gpt2 \\
& \texttt{--dpp\_scale\_factor} & 0.1 \\
\bottomrule
\end{tabular}%
}
\end{table*}

\begin{table}[t]
\centering
\caption{\textbf{Tuned \texttt{dbscan\_q} values.} This parameter controls the density quantile for clustering. It is tuned separately for each backbone and retriever combination to ensure the granularity of clusters matches the retriever's selection logic.}
\label{tab:params_dbscan_q}
\resizebox{\columnwidth}{!}{%
\begin{tabular}{l|l|ccc}
\toprule
\textbf{Backbone} & \textbf{Retriever} & \textbf{Banking77} & \textbf{CLINC150} & \textbf{HWU64} \\
\midrule
\multirow{3}{*}{\shortstack[l]{Qwen2.5\\7B-it}}
 & UCS+MDL   & 0.15 & 0.01 & 0.15 \\
 & UCS+DPP   & 0.01 & 0.01 & 0.01 \\
 & UCS+VoteK & 0.01 & 0.01 & 0.01 \\
\midrule
\multirow{3}{*}{\shortstack[l]{Llama-3.2\\3B-it}}
 & UCS+MDL   & 0.01 & 0.01 & 0.01 \\
 & UCS+DPP   & 0.01 & 0.01 & 0.01 \\
 & UCS+VoteK & 0.01 & 0.01 & 0.01 \\
\midrule
\multirow{3}{*}{\shortstack[l]{Gemma-2\\9B-it}}
 & UCS+MDL   & 0.01 & 0.01 & 0.15 \\
 & UCS+DPP   & 0.01 & 0.10 & 0.01 \\
 & UCS+VoteK & 0.01 & 0.01 & 0.5 \\
\bottomrule
\end{tabular}%
}
\end{table}

\begin{table}[!]
\centering
\caption{\textbf{Tuned \texttt{sgt\_lambda} values.} This parameter controls the strength of the UCS term ($\lambda$) added to the Retriever score. It balances the exploitation of the base method (MDL/DPP/VoteK) with the exploration of unseen clusters.}
\label{tab:params_sgt_lambda}
\resizebox{\columnwidth}{!}{%
\begin{tabular}{l|l|ccc}
\toprule
\textbf{Backbone} & \textbf{Retriever} & \textbf{Banking77} & \textbf{CLINC150} & \textbf{HWU64} \\
\midrule
\multirow{3}{*}{\shortstack[l]{Qwen2.5\\7B-it}}
 & UCS+MDL   & 0.01 & 0.005 & 0.001 \\
 & UCS+DPP   & 0.001 & 0.01 & 0.01 \\
 & UCS+VoteK & 5.0 & 5.0 & 5.0 \\
\midrule
\multirow{3}{*}{\shortstack[l]{Llama-3.2\\3B-it}}
 & UCS+MDL   & 0.05 & 0.05 & 0.05 \\
 & UCS+DPP   & 0.50 & 0.05 & 0.05 \\
 & UCS+VoteK & 0.05 & 0.05 & 0.50 \\
\midrule
\multirow{3}{*}{\shortstack[l]{Gemma-2\\9B-it}}
 & UCS+MDL   & 0.05 & 0.05 & 0.05 \\
 & UCS+DPP   & 0.50 & 0.10 & 0.05 \\
 & UCS+VoteK & 0.50 & 0.10 & 0.05 \\
\bottomrule
\end{tabular}%
}
\end{table}

\end{document}